\pdfoutput=1
\documentclass[11pt]{article}

\usepackage[]{acl}

\usepackage{times}
\usepackage{latexsym}
\usepackage{amsfonts}
\usepackage{amssymb}
\usepackage{pdfpages}
\usepackage{tabularx}
\usepackage{amsmath,bm}

\usepackage[T1]{fontenc}
\usepackage[utf8]{inputenc}

\usepackage{microtype}
\usepackage{amsmath}
\usepackage{diagbox}
\usepackage{graphicx} % Required for inserting images
\usepackage{cite}
\usepackage{booktabs}
\usepackage{multirow}
\usepackage{subcaption}
\usepackage{xurl}

\title{Question-Based Retrieval using Atomic Units for Enterprise RAG}

\author{Vatsal Raina, Mark Gales \\
  ALTA Institute, University of Cambridge \\
  \texttt{\{vr311,mjfg\}@cam.ac.uk} \\
  }

\begin{document}
\maketitle

\begin{abstract}

Enterprise retrieval augmented generation (RAG) offers a highly flexible framework for combining powerful large language models (LLMs) with internal, possibly temporally changing, documents. In RAG, documents are first chunked. Relevant chunks are then retrieved for a user query, which are passed as context to a synthesizer LLM to generate the query response. However, the retrieval step can limit performance, as incorrect chunks can lead the synthesizer LLM to generate a false response. This work applies a zero-shot adaptation of standard dense retrieval steps for more accurate chunk recall. Specifically, a chunk is first decomposed into atomic statements.  A set of synthetic questions are then generated on these atoms (with the chunk as the context). Dense retrieval involves finding the closest set of synthetic questions, and associated chunks, to the user query.  It is found that retrieval with the atoms leads to higher recall than retrieval with chunks. Further performance gain is observed with retrieval using the synthetic questions generated over the atoms. Higher recall at the retrieval step enables higher performance of the enterprise LLM using the RAG pipeline.

\end{abstract}

\section{Introduction}

Since the popularized ChatGPT as an instruction-finetuned large language model (LLM) deployed at scale to the lay market, there has been a substantial uptake on the interest of businesses to incorporate LLMs in their products for a variety of downstream tasks \citep{bahrini2023chatgpt, castelvecchi2023open, badini2023assessing, kim2024rag}.
For most companies, they are interested in using such models as enterprise LLMs where the model can handle queries related to proprietary on-premise data.

It has been repeatedly demonstrated that these LLMs have general (public) knowledge implicitly embedded in their parametric memory which can be extracted upon querying \citep{yu2023kola}. 
However, the LLMs do not have implicit knowledge about a specific enterprise's textual database in a custom domain and hence are prone to hallucinate in such situations \citep{xu2024hallucination, yu2023chain}.
Additionally, the transformer-based \citep{vaswani2017attention} LLMs typically have a limited context window (due to quadratic order in cost of the attention mechanism), which means information for a specific company to be queried over cannot be directly fed-in as a prompt to the LLM.
Due to limited budget, it is typically not feasible to finetune LLMs on a specific enterprise's data. In particular, with evolving data from ongoing projects, it is challenging to maintain a constantly updated company-specific LLM finetuned on new data without catastrophic forgetting \citep{luo2023empirical}.

To tackle this issue, and with retrieval augmented generation (RAG) proposed by \citet{lewis2020retrieval}, RAG-inspired systems have rapidly become the de-facto as a zero-shot solution for enterprise LLMs.
At the essence, there are 2 steps: 1. retrieval and 2. synthesis. Documents are split into independent chunks, and a retrieval process is applied to identify the relevant chunks to a given query. The retrieved chunks (which should fit into the context window) with the query are passed as the prompt to the synthesizer LLM to get the desired response.

Currently, the bottleneck for most enterprise LLMs is the retrieval step, where the correct information is not retrieved for the LLM to answer the question \citep{arora2023gar}. 
Hence, this work focuses on building upon zero-shot approaches to improve the retrieval step for RAG. 
A potential limitation of the RAG set-up is that an embedding model is used to retrieve the relevant chunks efficiently when given a query. Each pre-calculated chunk has its corresponding embedding stored in memory, which allows the closes chunks to be retrieved by embedding the incoming query into the same space. However, there is a mismatch in trying to match the space of queries and chunks as each chunk can carry a large amount of information.

Instead, our work looks to represent each chunk as a set of atomic pieces of information. \citet{min2023factscore} introduced atomization of text for improving the assessment of summary consistency.  These atoms can be structural (e.g. sentences of a chunk) or unstructured where a set of atoms is generated for any chunk. By embedding the atoms instead of the chunks themselves, the relevant atoms can instead be identified (that correspond to a specific set of chunks) for the posed query in the embedding space. The atomic breakdown of the chunk enables more accurate retrieval. 

We further identify that even with the atomic embedding representations of the chunk, a given atom and the query do not necessarily best align for retrieval as the former is a statement with a piece of information while the latter is a question about locating a missing piece of information. Thus, we propose generating synthetic atomic questions. Each atom has a set of questions generated, which in turn are embedded. Therefore, the embedded incoming query is used to identify the closest set of atomic questions which in turn point to the relevant set of chunks to be passed to the synthesizer LLM in the RAG pipeline.
As enterprise RAG operates over a closed set of documents, the generation of the atoms and corresponding synthetic questions is a one-off cost. Similarly, the increased set of embeddings to search over for the closest matches for the query embedding is of less concern given the various very efficient algorithms for embedding search such as FAISS \citep{douze2024faiss}. 

Current information retrieval approaches applied to the RAG pipeline look at improving the quality of dense retrieval through generation augmented retrieval (GAR), where a query is rewritten for high recall retrieval. However, we focus our attention on representing the chunks more efficiently for retrieval (information retrieval literature explore such approaches - see Section \ref{sec:related}.
The contributions: an exploration of how the retrieval step in the enterprise RAG pipeline is improved with structured and unstructured atomic representation of a document chunk and further improvement with the generation of atomic questions.

\section{Related Work}
\label{sec:related}

Recently, several works have extended RAG \citep{zhao2024retrieval}. Many approaches finetune the components of the RAG pipeline. For example, \citet{siriwardhana2023improving} explore adapting end-to-end RAG systems for open-domain question-answering while \citet{zhang2024raft} introduce RAFT for finetuning RAG systems on specific domains by learning to exclude distractor documents. Additionally, \citet{siriwardhana2021fine,lin2023ra} jointly train the retriever and the generator for target domains. 
However, our work focuses on exploring zero-shot solutions as finetuning can be a computationally infeasible procedure for many enterprises.

In terms of zero-shot approaches, there have been several extensions proposed. \citet{gao2023precise} propose hypothetical document embedding (HyDE) where an LLM is used to transform the input query into an answer form (hallucinations are acceptable) for improved dense retrieval over the chunks. Similarly, \citet{wang2023query2doc} suggest a query expansion approach termed query2doc where an LLM is used to expand the query \citep{jagerman2023query} with a pseudo-generated document, which they demonstrate to be effective for dense retrieval. 
Alternatively, we propose approaches that focus on modifying the knowledge base on which retrieval is performed rather than modifying the user queries as is common in GAR \citep{shen2023large, feng2023knowledge, arora2023gar}.

\citet{song2024re3val} retrieve a superfluous number of chunks during the retrieval step. They then re-rank the retrieved chunks with a re-ranker system to identify the most relevant set. Similarly, \citet{wang2023learning} propose FILCO to filter out the retrieved documents as an additional step in the RAG pipeline. \citet{sun2023chatgpt} explore the zero-shot use of LLMs as alternatives for traditional re-rankers. \citet{arora2023gar} additionally incorporate the re-rank steps with GAR in an iterative feedback loop. Leveraging the comparative abilities of LLMs, \citet{qin2023large} propose using pairwise comparisons for the re-ranking of retrieved documents. Alternatively, \citet{sarthi2024raptor} propose RAPTOR as an iterative technique to pass a summarized context (based on the retrieved documents) to the synthesizer. Iter-RetGen by \citet{shao2023enhancing} follow a similar iterative summarization strategy with LLMs. Finally, ActiveRAG \citep{xu2024activerag} encourages the synthesizer to consider parametric memory rather than just relying on the set of retrieved documents.
% Re-ranking and summarization approaches are viewed as adjunct post-processing steps that are complementary with the atomic retrieval approaches proposed in this work.
\citet{gao2023retrieval} summarize all advanced RAG approaches as additional pre-retrieval or post-retrieval steps. Pre-retrieval steps include query routing, query re-writing and query expansion. Post-retrieval steps include re-ranking summarization and fusion. The synthetic question retrieval over atomized units from the document set is a form of pre-retrieval that operates on the knowledge store rather than on the user query. Hence, our work remains complementary with all forms of post-retrieval RAG. See Appendix Figure \ref{fig:adapted_rag_summary}.

Traditionally, retrieval of relevant documents for a given query has been well studied \citep{hambarde2023information} with approaches such as BM25 \citep{robertson2009probabilistic}. In recent years, dense retrieval approaches have dominated as efficient retrieval processes where queries and documents are represented as dense vectors (embeddings) and documents are retrieved based on the similarity between these vectors. Semantically meaningful vectors have been possible with the series of regularly updated sentence transformers for generating general purpose embeddings including Sentence-BERT \citep{reimers2019sentence}, ConSERT \citep{yan2021consert}, SimCSE \citep{gao2021simcse}, DiffCSE \citep{chuang2022diffcse}, sentence-T5 \citep{ni2022sentence} and E5 \citep{wang2022text}. More recently, there have been a series of more powerful embedding models that adapt instruction-finetuned language models as embedders \citep{li2023towards, SFRAIResearch2024, muennighoff2024generative, wang2023improving, behnamghader2024llm2vec}. Therefore, this work restricts exploration to dense retrieval.

In recent information retrieval literature, \citet{chen2023dense} explore what granularity should be used for retrieval. They introduce the concept of breaking a passage into atomic expressions where each encapsulates a single factoid. \citet{zhang2022multi} argue a document consists of many diverse details. Hence, they propose representing a document using multiple (diverse) embeddings to capture different views of the same content.
\citet{gospodinov2023doc2query} investigate for Doc2Query, a method of expanding the content of a document, how hallucinations can be minimized in the generated queries over a document.
Our work connects these concepts for specifically generating multiple synthetic questions over atoms in enterprise RAG.
This work is a bridge between methods explored in information retrieval and the RAG community.

\section{Retrieval for RAG}

\begin{figure*}[t]
     \centering
\includegraphics[width=2.0\columnwidth]{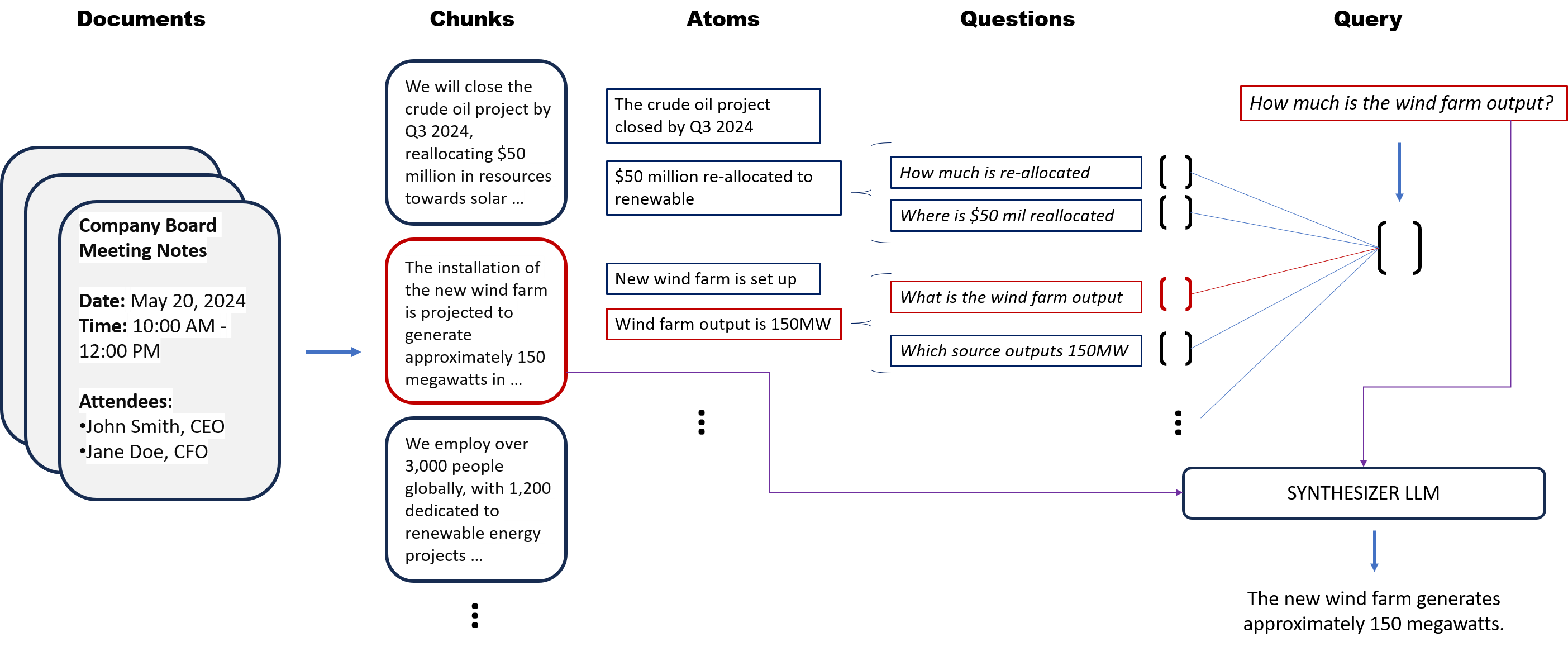}
        \caption{Question-based retrieval using atomic units for enterprise RAG.}
        \label{fig:atomicrag}
\end{figure*}

In enterprise RAG systems, the core pipeline can be summarized as follows.
\begin{enumerate}
    \item \textbf{Split}: Given a textual corpus of documents, a set of chunks are generated by splitting all text into distinct paragraphs.
    \item \textbf{Retrieve}: For a given user query, the relevant set of chunks are retrieved.
    \item \textbf{Synthesize}: The original query and the retrieved chunks are passed to a synthesis model to generate a response to the query using the provided chunk information as the context.
\end{enumerate}
Here, the focus is on improving the retrieval step of the enterprise RAG pipeline. For the scope of the data considered in this work, we assume that the answer to a specific query is present in only one chunk (i.e. there are no unanswerable queries and multiple chunks are not required to deduce the answer to a question). Therefore, the retrieval step task can be defined as follows:
\newline\newline
\noindent
\textbf{Task} Let $R(q;c) \in {0,1}$ denote an oracle relevancy function that returns 1 if a chunk, $c$, contains the answer to the user query $q$ and 0 otherwise. Given a set of $N$ chunks, $\{c\}_{1:N}$, and a user query $q$, retrieve chunk $c_k$ such that $R(q;c_k) = 1$ but $ \sum_{i\neq k} R(q;c_i) = 0$.
\newline\newline
\noindent  
Next, we describe the various approaches for the retrieval step of enterprise RAG systems. The focus is on zero-shot approaches that can be applied without any training and we assume we have no-cost in accessing the relevancy function.

\subsection{Standard}

In the standard retrieval set-up for the RAG pipeline, dense retrieval is used for identifying the most relevant chunk to the user query. Let $E\left(\cdot\right)$ denote a sentence embedding model. The embedding model has been trained to produce semantically meaningful vector representations of natural language text (see Section \ref{sec:related} for the evolution of sentence transformers). All of the document chunks and the query are embedded into the high-dimensional space such that:
\begin{equation}
    \mathbf{c}_i = E\left(c_i\right), \forall i \in [1,N] 
\end{equation}
\begin{equation}
    \mathbf{q} = E(q)
\end{equation}
Then the chunk, $c_{\hat{k}}$, is selected such that $\mathbf{c}_{\hat{k}}$ and $\mathbf{q}$ have the shortest cosine distance between all chunk embeddings and the query embedding. The cosine distance between a pair of vectors $\mathbf{a}$ and $\mathbf{b}$ is defined as $\texttt{cos}[\mathbf{a},\mathbf{b}] = 1 - \mathbf{a}^T\mathbf{b} / |\mathbf{a}||\mathbf{b}|$.
\begin{equation}
\label{eq:standardrag}
  [\texttt{chunk}] \hspace{3mm}   \hat{k} = \arg\min_{k} \texttt{cos}[\mathbf{q},\mathbf{c}_{k}]
\end{equation}

\noindent One shortcoming of the standard retrieval approach in RAG is that query embeddings are compared against chunk embeddings. However, the semantic embedding representation of a query does not necessarily align with the semantic embedding representation of the chunk that needs to be retrieved. Hence, dense retrieval can lead to the incorrect chunk being retrieved. The following sections describe modifications to the dense retrieval of the chunks to increase the recall rate.

\subsection{Generation augmented retrieval}

As a baseline, the HyDE approach \citep{gao2023precise} is used as a form of GAR (see Section \ref{sec:related}) \footnote{There are several GAR approaches. We find the form of HyDE works best for this dataset from preliminary experiments and hence select it as an appropriate baseline for GAR in RAG.}. The approach requires the query, $q$ to be re-written to $q'$ where $q'$ aims to be a complete hypothesized answer to the query. For example, \textit{What is the capital of India?} is rewritten to \textit{The capital of India is London}. Note, the answer of the query is not important. Instead the form of the answer should hopefully match the nature of the real answer e.g. \textit{London} and \textit{New Delhi} are both places. 
Now, the standard retrieval approach is applied from Equation \ref{eq:standardrag} with $\mathbf{q'} = E(q')$ as the embedding of the re-written query.

\begin{equation}
\label{eq:garrag}
  [\texttt{hyde}] \hspace{3mm}   \hat{k} = \arg\min_{k} \texttt{cos}[\mathbf{q'},\mathbf{c}_{k}]
\end{equation}
Intuitively, with an answer-like sequence present in the embedded query, there is a greater likelihood of matching with the relevant chunk. Typically, the re-writing process is achieved zero-shot with an LLM by relying on its parametric answer (at the rewriting stage, hallucinations are not a concern). Henceforth, this approach is referred to as HyDE.

\subsection{Atomic}
\label{sec:theoryatomic}

A query is typically searching for a specific piece of information in a chunk. The embedding representation of the chunk can be viewed as an average representation of all the different pieces of information present in the chunk. Often, the pieces of information in the same chunk can be distinct, which can lead to the query embedding being distant from the target chunk embedding with the answer. 

Therefore, we explore atomic retrieval. Here, the chunk text is partitioned into a set of atomic statements (referred henceforth as atoms) such that 
\begin{equation}
    c_k \rightarrow \{ a_1^{(k)}, \hdots a_{n_k}^{(k)} \}, \forall k
\end{equation}

\noindent 
With $\mathbf{a} = E(a)$, the query embedding is compared against the atomic embeddings. The closest atomic embedding is used to identify the corresponding chunk to be retrieved. The expectation is that individual atomic embeddings are more likely to align with a query's embedding in the vector space.
\begin{equation}
\label{eq:atom}
      [\texttt{atom}] \hspace{3mm}  \hat{k}, \hat{j} = \arg\min_{k,j} \texttt{cos}[\mathbf{q},\mathbf{a}_j^{(k)}]
\end{equation}

\noindent For evaluation, $\hat{k}$ is of interest and $\hat{j}$ is discarded. In this work two forms of atoms are considered:
\begin{itemize}
    \item \textbf{Structured}: Each sentence in the chunk is a separate atom. 
    \item \textbf{Unstructured}: An atom generation system is asked to generate atomic statements that best capture all the information in the chunk. See Section \ref{sec:model} for a description of the specific atom generation system.
\end{itemize}

\noindent Despite atomizing a chunk of text, there is risk of the query not necessarily matching the target atom in the embedding space as the atom contains semantic information about the answer while the query does not. Therefore, we propose an extension called atomic questions. For a given atom, a set of synthetic questions are generated that are best answered by the atom given the chunk as the context information. Hence,
\begin{equation}
    a_j^{(k)} \rightarrow \{ y_1^{(j,k)}, \hdots y_{n_{j,k}}^{(j,k)} \}, \forall j,k
\end{equation}
\begin{equation}
\label{eq:question}
      [\texttt{question}] \hspace{3mm}  \hat{k}, \hat{j}, \hat{i} = \arg\min_{k,j,i} \texttt{cos}[\mathbf{q},\mathbf{y}_i^{(k,j)}]
\end{equation}
As before, only $\hat{k}$ is of interest for evaluation. Figure \ref{fig:atomicrag} summarizes the RAG pipeline with question-based retrieval using atomic units. Effectively, each chunk can be summarized by a set of questions that probe different pieces of information.

\section{Experiments}

\subsection{Data}

\begin{table}[htbp!]
\centering
\small 
\begin{tabular}{l|cc}
\toprule
& SQuAD & BiPaR \\
\midrule  
\# total chunks & $2,067$ & $375$  \\
\# total queries & $10,570$ & $1,500$ \\
\# queries / chunk  & $5.1_{\pm 2.3}$ & $4.0_{\pm 0.0}$ \\
\# words / query & $10.2_{\pm 3.6}$ & $7.2_{\pm 2.9}$ \\
\# words / chunk  & $122.8_{\pm 54.8}$ & $181.1_{\pm 52.8}$ \\
\# sentences / chunk & $6.6_{\pm 3.1}$ & $14.2_{\pm 5.7}$ \\
   \bottomrule
    \end{tabular}
\caption{Statistics of datasets.}
\label{tab:data}
\end{table}

\noindent SQuAD \citep{rajpurkar2016squad} is a popular choice as an extractive reading comprehension dataset consisting of triples of contexts, questions and answer extracts. The contexts are sourced across a wide variety of Wikipedia articles. We re-structure the validation split of the SQuAD dataset for the task of retrieval in RAG as follows. As all questions are answerable (unlike SQuAD 2.0 \citep{rajpurkar2018know}), we assume that the answer to a given question must be present in its corresponding context passage. We additionally assume that the answer to a specific question is not present in any other context. Therefore, we shuffle all the contexts such that the task requires retrieval of the appropriate context for a given question. Once a particular context is retrieved, it is the role of the synthesizer in the RAG pipeline to generate the required answer. Remaining consistent with the terminology of retrieval in RAG, contexts are viewed as chunks and the questions are termed queries. The collection of chunks are effectively the pre-split texts from a knowledge store, which in this case is Wikipedia.

Table \ref{tab:data} summarizes the statistics of the re-structured SQuAD validation set for assessing the RAG framework. In total there are 2,067 chunks with 10,570 queries, resulting in approximately 5 queries per chunk. The number of sentences within each chunk vary with a single standard deviation of 3.1 about 6.6. As mentioned, in Section \ref{sec:theoryatomic}, the sentences of a chunk are treated as structured atoms. Overall, the re-structured dataset allows us to explore whether we can improve the retrieval of chunks for queries over a fixed knowledge store.
% by adapting the manner in which the knowledge store is represented.

Additionally, we consider BiPaR \citep{jing2019bipar} for evaluating the RAG framework. 
BiPaR is a manually annotated dataset of bilingual parallel texts in a novel-like style, created to facilitate monolingual, multilingual, and cross-lingual reading comprehension tasks. We focus on only the English texts over the test split.
In a similar vain to SQuAD, the knowledge store is constructed by shuffling the contexts for all queries. Table \ref{tab:data} summarizes the main details. It is particularly useful to consider BiPaR for enterprise RAG as the information content of the context is based on extracts from novels. As the stories are fictional and not factual, the parametric memory of an LLM cannot expect to know the answers to the queries. Therefore, BiPaR mimics the set-up of proprietary knowledge stores for enterprises where retrieval is necessary to identify the relevant information for a query.

\subsection{Model details}
\label{sec:model}

\begin{table}[htbp!]
\centering
\small 
\begin{tabular}{p{1.8cm}|p{5cm}}
\toprule
Task & Prompt \\
\midrule
Query re-writing & Please write a full sentence answer to the following question. \{query\} \\
\midrule
Unstructured atom generation &
Please breakdown the following paragraph into stand-alone atomic facts. Return each fact on a new line. \{chunk\}
\\
\midrule
Question generation & 
Generate a single closed-answer question using: \{chunk\}
The answer should be present in: \{atom\}
\\
   \bottomrule
    \end{tabular}
\caption{ChatGPT prompts for zero-shot tasks.}
\label{tab:prompts}
\end{table}

\noindent For generating the embedding representations, the embedder $E(\cdot)$ is selected as all-mpnet-base-v2 \footnote{\url{https://huggingface.co/sentence-transformers/all-mpnet-base-v2}} from Huggingface. This embedder is a popular choice for enterprise RAG (the default in LlamaIndex \footnote{\url{https://www.llamaindex.ai/}} for open-source LLMs) as it performs well on the MTEB \citep{muennighoff2023mteb} leaderboard despite its small size of 110M parameters. We additionally present results using the e5-base-v2 \footnote{\url{https://huggingface.co/intfloat/e5-base-v2}} embedder \citep{wang2022text}, which has topped the MTEB leaderboard for models of the base size.

Instruction-tuned LLMs \citep{touvron2023llama, jiang2023mistral} have demonstrated impressive capabilities across a diverse range of tasks. Therefore, for HyDE, the query re-writing process is achieved with zero-shot usage of ChatGPT 3.5 Turbo \footnote{\url{https://platform.openai.com/docs/models}}. Similarly, ChatGPT is used for generating atomic statements from a chunk of text as described in Section \ref{sec:theoryatomic}. Finally, we make use of the same model to automatically generate questions on the atoms. Table \ref{tab:prompts} summarizes the prompts for each of these tasks \footnote{Manual prompt engineering was performed to identify the appropriate prompts to achieve sensible results.}. The question generation system is applied for a maximum of 15 times on each atom \footnote{The code will be made available if accepted.} at which the performance plateaus (see Section \ref{sec:results}).

\subsection{Evaluation}

In information retrieval, there is a large number of metrics proposed for assessing retrieval capabilities \citep{arora2016evaluation}. Here, we focus on calculating R@K (recall at $K$). R@K calculates the fraction of queries for which the correct chunk is within the top K chunks when retrieval is performed. We specifically present R@1, R@2 and R@5. Note, R@1 checks for the exact match while R@2 and R@5 are more lenient. We do not consider other retrieval measures that account for the ordering of the documents retrieved as in the scope of this work there is only 1 relevant chunk for each query.
For RAG, it is of interest to return multiple chunks from the retrieval step and leave the job of finding the correct answer amongst the retrieved chunks to the synthesizer. The limit on this approach is the context window of the synthesizer. For example the context window for ChatGPT 3.5 is 16K tokens. Hence, we consider moderately high K for R@K.

\begin{table*}[t]
\centering
\small 
\begin{tabular}{lll|ccc|ccc}
\toprule
 \multirow{2}{*}{Dataset} & \multirow{2}{*}{Item} & & \multicolumn{3}{c|}{all-mpnet-base-v2} & \multicolumn{3}{c}{e5-base-v2} \\
 &  &  & R@1 & R@2 & R@5 & R@1 & R@2 & R@5 \\
\midrule
\multirow{8}{*}{SQuAD} & \multirow{2}{*}{Chunk} & Text & 65.5 & 78.9 & 89.3 & 76.2 & 87.1 & 94.4 \\
& & HyDE & 65.2 & 77.9 & 88.9 & 66.4 & 79.9 & 91.1 \\
\cmidrule{2-9}
& \multirow{3}{*}{Atom-Structured} & Text & 70.2 & 81.4 & 90.6 & \underline{80.1} & \textbf{89.3} & \textbf{95.1} \\
& & HyDE & 71.5 & 82.3 & 91.1 & 73.7 & 84.6 & 93.0  \\
& & Question & \underline{73.8} & 83.5 & 91.2 & 78.1 & 87.2 & 93.8 \\
\cmidrule{2-9}
& \multirow{3}{*}{Atom-Unstructured} & Text & 72.6 & \underline{83.9} & \underline{91.9} & 80.0 & 88.3 & \underline{94.6} \\
& & HyDE & 73.1 & 83.7 & 91.7 & 73.9 & 84.4 & 92.3 \\
& & Question  & \textbf{76.3} & \textbf{85.4} & \textbf{92.6} & \textbf{80.2} & \underline{88.6} & 94.5 \\
\midrule\midrule 
\multirow{8}{*}{BiPaR} & \multirow{2}{*}{Chunk} & Text & 33.7 & 43.1 & 54.7 & 42.1 & 52.6 & 63.7 \\
& & HyDE & 31.2 & 41.2 & 51.7 & 36.6 & 47.4 & 58.9 \\
\cmidrule{2-9}
& \multirow{3}{*}{Atom-Structured} & Text & 42.6 & 52.3 & 65.4 & 47.7 & 57.8 & 69.5 \\
& & HyDE & 40.1 & 50.1 & 62.1 & 43.5 & 52.1 & 64.9  \\
& & Question & \textbf{53.8} & \textbf{63.4} & \textbf{73.3} & \textbf{55.9} & \textbf{64.8} & \textbf{75.3} \\
\cmidrule{2-9}
& \multirow{3}{*}{Atom-Unstructured} & Text & 43.9 & 54.3 & 66.9 & 49.7 & 58.1 & 69.1 \\
& & HyDE & 41.7 & 52.5 & 64.6 & 43.0 & 51.7 & 63.7 \\
& & Question  & \underline{53.7} & \underline{61.9} & \underline{72.9} & \underline{55.3} & \underline{64.1} & \underline{74.5}  \\
   \bottomrule
    \end{tabular}
\caption{Retrieval performance for enterprise RAG. All recall rates are represented as percentages.}
\label{tab:main_results}
\end{table*}

\section{Results}
\label{sec:results}

Table \ref{tab:main_results} presents the recall rates with various zero-shot approaches of the retrieval step using SQuAD and BiPaR with 2 different embedders.

Let's take a look first at the all-mpnet-base-v2 embedder for SQuAD. Operating at the chunk scale, where the raw text is embedded for dense retrieval, the standard RAG achieves a recall of 65.5\% with the top 1, which increases to 89.3\% when considering the top 5 chunks retrieved. By applying GAR with the HyDE baseline at the chunk scale, we do not observe gains. As discussed in Section \ref{sec:theoryatomic}, the text chunk contains several semantic pieces of information while the re-written query remains related to a single semantic piece of information. Hence, it is challenging for the HyDE approach to improve recall at the chunk scale.

By splitting a chunk into structured atoms (sentences), Table \ref{tab:main_results} further shows the recall by embedding the atomic text or the corresponding synthetic questions generated on those atoms (Equations \ref{eq:atom} and \ref{eq:question} respectively).
Additionally, the HyDE approach is applied with the atomic embeddings, using the rewritten query instead of the original from Equation \ref{eq:atom}.
Embedding the atomic text instead of the chunk text observes significant gains, reaching 70.2\% for R@1 and 90.6\% for R@5. As the length of a sentence in a chunk is closer in length to the re-written query, the HyDE approach on the structured atoms further boosts the recall rates. An additional gain is again observed by performing dense retrieval with the set of generated questions, achieving up to 73.8\% for R@1.

The final rows of Table \ref{tab:main_results} for SQuAD with all-mpnet-base-v2 further demonstrates the benefits of using unstructured atoms in place of the structured atoms. A sentence from a chunk contains more granular information than the whole chunk but is not necessarily constrained to one piece of atomic information. Therefore, by re-writing the chunk into a series of independent atoms, dense retrieval between the query and the set of atomic embeddings leads to higher recall rates. As with the structured atoms, the HyDE approach leads to further performance gains with the unstructured atoms. Finally, we observe the best performance across all three recall rates by applying dense retrieval using the generated questions on the atoms. It is clear that higher recall retrieval is possible by matching queries with questions as they can expect to be of the same form rather than attempting to match queries with chunks.

\begin{figure*}[t]
    \centering
    \begin{tabular}{cc}
        \begin{subfigure}[b]{0.48\textwidth}
            \centering
            \includegraphics[width=\textwidth]{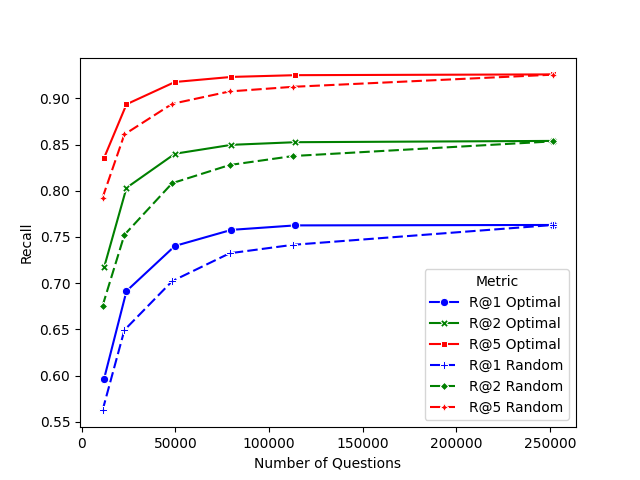}
            \caption{SQuAD: all-mpnet-base-v2}
            \label{fig:subfig1}
        \end{subfigure} &
        \begin{subfigure}[b]{0.48\textwidth}
            \centering
            \includegraphics[width=\textwidth]{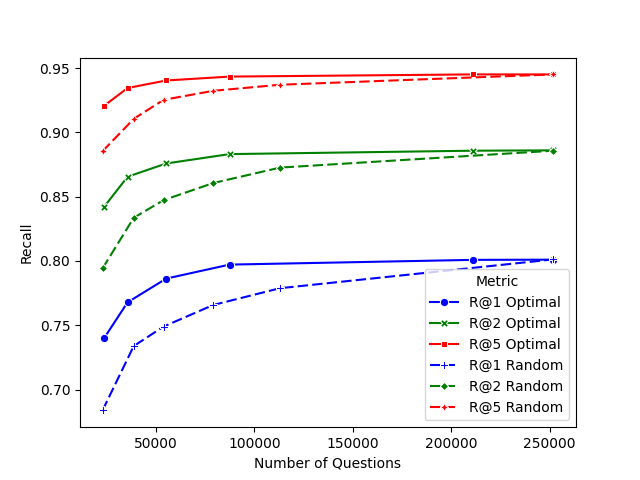}
            \caption{SQuAD: e5-base-v2}
            \label{fig:subfig2}
        \end{subfigure} \\

        \begin{subfigure}[b]{0.48\textwidth}
            \centering
            \includegraphics[width=\textwidth]{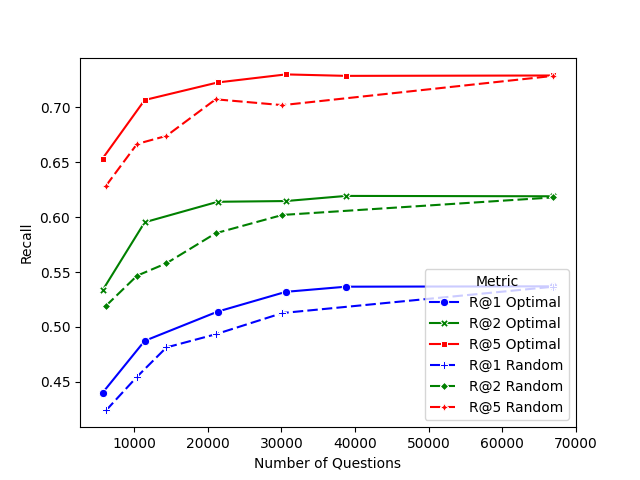}
            \caption{BiPaR: all-mpnet-base-v2}
            \label{fig:subfig3}
        \end{subfigure} &
        \begin{subfigure}[b]{0.48\textwidth}
            \centering
            \includegraphics[width=\textwidth]{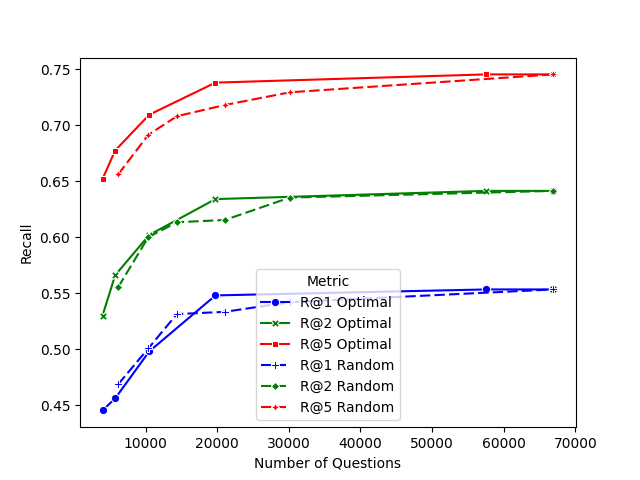}
            \caption{BiPaR: e5-base-v2}
            \label{fig:subfig4}
        \end{subfigure}
    \end{tabular}
    \caption{Efficient unstructured atomic question retrieval. See Appendix Figure \ref{fig:efficiency_optimal}, Section \ref{sec:app:open} for additional models.}
    \label{fig:efficiency}
\end{figure*}

Considering the higher performing embedder e5-base-v2 on SQuAD, the trends are less clear due to a stronger baseline. We observe that for R@1 that atomic question retrieval with unstructured atoms has the best performance, but drops to second and third highest for R@2 and R@5 respectively.

Let's now consider BiPaR from Table \ref{tab:main_results}. Very similar trends are observed for both all-mpnet-base-v2 and e5-base-v2 embedders on this dataset. It is noticeable that HyDE at both the chunk, structured atoms and unstructured atoms struggles to outperform the equivalent text. This deviation in the trend observed in SQuAD is expected as BiPaR is based on fictional stories while SQuAD is based on factual Wikipedia articles. Hence, the hallucinated answers generated by HyDE are unlikely to help with retrieving relevant chunks which do not correspond to the re-written query (see Appendix Section \ref{sec:app:hyde} for more analysis about HyDE). In contrast, for public factual information (as in SQuAD), the hypothesized answer generated by a powerful LLM is more likely to be the correct answer than a hallucination. Question-based retrieval operating on atoms demonstrates significant gains over the baseline for BiPaR. For example, using e5-base-v2 improves R@1 by approximately 14\%.

In general, for the re-formatted SQuAD dataset, Table \ref{tab:data} states there are 2,067 unique chunks. Therefore, the standard retrieval approach for RAG leads to storing 2,067 chunk embeddings. In contrast, the atomic retrieval has substantially larger number of embeddings stored. Using structured atoms, there are 13,630 sentences in total while there are 16,793 unstructured atoms across the corpus. 
By considering the synthetic question generation strategy described in Section \ref{sec:model}, question retrieval strategies require $13,630\times 15$ and $16,793\times 15$ embeddings to be stored in memory for structured atoms and unstructured atoms respectively. A similar increase in the storage of embeddings apply for the BiPaR dataset. Hence, it is of interest to explore how the number of questions required for each atom can be reduced to remove the redundant ones.

Figure \ref{fig:efficiency} presents how the performance varies with the number of synthetically generated questions on the unstructured atoms. For each recall rate (R@1, R@2 and R@5), two profiles are indicated: 1. a random selection of synthetic questions for the atoms of each chunk; 2. an optimally diverse selection of synthetic questions for the atoms of each chunk. The optimally diverse set of questions is selected as follows. A threshold, $\tau$ is selected on the pairwise cosine distance. For the full set of atomic questions generated, the pairwise cosine distances of the question embeddings is calculated for each chunk. If any pairwise cosine distance is below $\tau$, one of the questions is purged. The process if repeated until all questions in the remaining set have pairwise cosine distances of their embeddings above $\tau$. By sweeping $\tau$, the total number of synthetic questions across the corpus changes.
One can hence expect that a chunk with more information will have a more diverse set of questions.  

 Figure \ref{fig:efficiency} shows that a significant number of questions are redundant across the SQuAD and BiPaR chunks. By removing more than half of the questions (and hence halving the storage cost), performance can be maintained at the maximal value for each of the recall rates. In the extreme setting, with only 20\% of the questions retained, there is only a marginal decrease in recall when using the optimal set. Thus, despite a larger storage cost with atomic question retrieval compared to standard enterprise RAG, the performance boost can be justified with an efficient choice of synthetic questions to retain. See Appendix Section \ref{sec:app:unans} for unanswerability analysis of the generated questions.

\section{Conclusions}

RAG systems are a popular framework for enterprises for automated querying over company documents. However, poor recall of relevant chunks with dense retrieval causes errors to propagate to the synthesizer LLM. Previous works have focused on extensions involving generation augmented retrieval where the query is re-written at inference time to improve recall. Conversely, we explore adaptations to the storage of the chunks.
The retrieval step for RAG can be refined in a zero-shot manner by 1) atomizing the chunks and 2) generating questions on the atoms. Significant improvements are observed on the BiPaR and SQuAD datasets with this approach as partitioning a chunk into atomic pieces of information allows dense retrieval with the query to be more effective. Moreover, operating in the question space, the query embedding aligns better with the synthetic questions of the target chunk. We further demonstrate that the storage cost of a large number of synthetic question embeddings can be dramatically reduced by only storing a diverse set of questions for each chunk. Question-based retrieval using atomic units will enable the deployment of higher performing enterprise RAG systems without relying on any additional training. 

% \section{Acknowledgements}
% This research is partially funded by the EPSRC (The Engineering
% and Physical Sciences Research Council)
% Doctoral Training Partnership (DTP) PhD studentship
% and supported by Cambridge University Press \& Assessment (CUP\&A), a
% department of The Chancellor, Masters, and Scholars
% of the University of Cambridge.

\section{Limitations}

In this work, we have made several assumptions which do not necessarily hold in real enterprises. Our work focuses on only closed queries where a single atom contains the answer. It would be interesting to extend the approach to handle multi-hop situations by generating synthetic questions on pairs or collections of atoms. Additionally, we have focused the presentation of our results on SQuAD and BiPaR. It will be useful to consider additional standard information retrieval benchmarks such as the BEIR datasets \citep{thakur2021beir}. We specifically focus on small-scale datasets due to limitations in the available computational budget. We do emphasise that small-scale datasets often mimic the size of datasets in enterprises, which emphasises our focus on enterprise RAG. We further emphasise that for the use case of enterprise RAG, queries are over proprietary information. Most mainstream information retrieval datasets are based on public factual information, which is not convincing for the enterprise set-up. BiPaR (our choice of dataset) is based on information from stories (non-factual), which is more aligned with the concept of proprietary information.

% \section{Future work}

% We will extend this work to additional datasets in different domains. In particular, we will include the results of this approach for the financial and the medical domains.

\section{Ethics statement}

There are no ethical concerns with this work.

\bibliographystyle{acl_natbib}
\bibliography{import}

\appendix

\begin{table*}[htbp!]
    \centering
    \begin{tabular}{l|ccccc}
        \toprule
        Model & nDCG@1 & nDCG@3 & nDCG@5 & nDCG@10 & R@10 \\
        \midrule
        BM25 & $18$ & $30$ & $35$ & $40$ & $67$ \\
        all-MiniLM-L6-v2 & $29$ & $43$ & $48$ & $53$ & $79$ \\
        BGE-base & $37$ & $54$ & $59$ & $61$ & $85$ \\
        E5-base-v2 & $41$ & $57$ & $61$ & $64$ & $87$ \\
        \midrule 
        E5-base-v2 (ours) & $36$ & $51$ & $54$ & $58$ & $82$ \\
        + HyDE & $42$ & $57$ & $61$ & $63$ & $85$ \\
        \midrule
        all-mpnet-base-v2 & $37$ & $51$ & $56$ & $61$ & $87$ \\
        + HyDE & $39$ & $56$ & $60$ & $63$ & $88$ \\
        \bottomrule
    \end{tabular}
    \caption{Baselines for ClapNQ with HyDE.}
    \label{tab:clapnq}
\end{table*}

\begin{figure*}[htbp!]
    \centering
            \includegraphics[width=\textwidth]{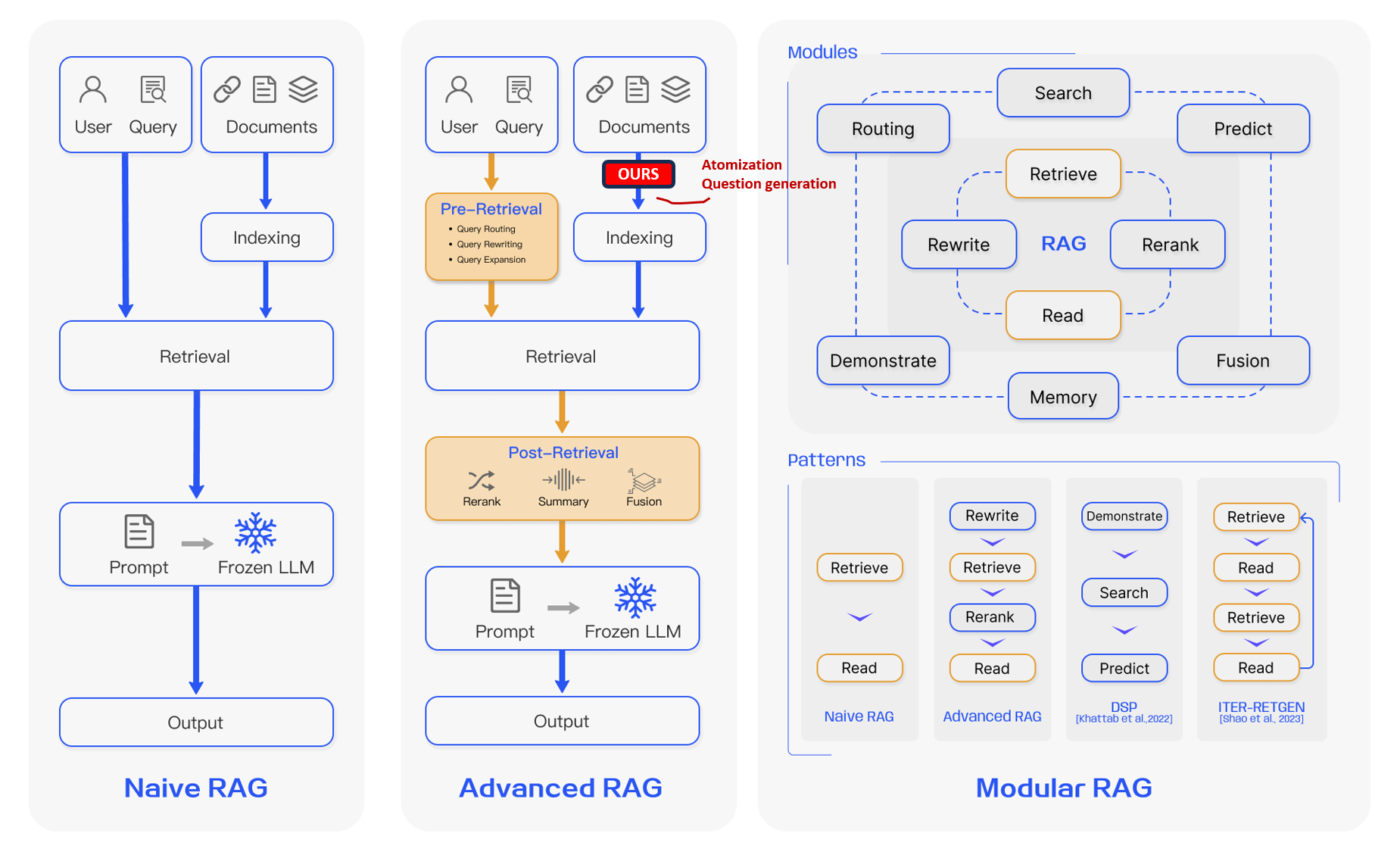}
    \caption{Adapted diagram from \citet{gao2023retrieval} to summarize existing RAG approaches. We highlight in \textit{red} our contribution to the advanced RAG panel. Specifically, we modify the documents before they are indexed using atomization and synthetic question generation.}
    \label{fig:adapted_rag_summary}
\end{figure*}

\newpage

\section{Drawback of HyDE}

\label{sec:app:hyde}

In the main paper, we observed that HyDE performs well for SQuAD but is less impressive for BiPaR. 
This Section aims to revisit how HyDE operates to explain the difference. Qualitatively, HyDE uses the parametric memory of an LLM to re-write the query as a complete sentence that answers the query. The re-written query is then used to retrieve the relevant chunks. The HyDE paper emphasizes that it doesn't matter if the answer is hallucinated as the form of the hypothetical answer can expect to be aligned with the chunk containing the correct answer.

However, it is clear that HyDE struggles on BiPaR while working well on SQuAD.
We suspect the reason for this discrepancy is that SQuAD is based on publicly known factual information from Wikipedia while BiPaR is based on fictional stories. Therefore, when HyDE is applied on SQuAD, the hypothesized answer often is simply the correct answer itself, leading to an artificial boost in the retrieval performance. The correct answer is generated typically by the parametric memory of a powerful LLM used for the query re-writing. In contrast, as the answers to the queries in BiPaR are not within the scope of general knowledge, the hypothesized answer from HyDE does not help in boosting the retrieval performance.

In order to investigate the dependence of HyDE on factual information for improving retrieval performance, we do additional analysis. We select CLAPNQ \citep{rosenthal2024clapnq} as a recently curated RAG dataset where the knowledge store is based on publicly available information (like SQuAD). Additionally, CLAPNQ has been exclusively designed for long-form answers. Therefore, we expect HyDE to demonstrate significant performance gains on this dataset as the hypothesized answer is likely to be the correct answer with high overlap with the target chunk due to the length of the answer. We show our results as follows in Table \ref{tab:clapnq}. The top 5 rows are quoted directly from \citet{rosenthal2024clapnq}. As well as recall, we report nDCG \citep{jarvelin2002cumulated} here as a standard retrieval metric used in \citet{rosenthal2024clapnq} where the order of the retrieved chunks is accounted for in calculating the performance. It is clear for both of our implementations that HyDE demonstrates retrieval performance gains on this challenging RAG dataset.

\begin{figure*}[t]
    \centering
    \begin{tabular}{cc}
        \begin{subfigure}[b]{0.33\textwidth}
            \centering
            \includegraphics[width=\textwidth]{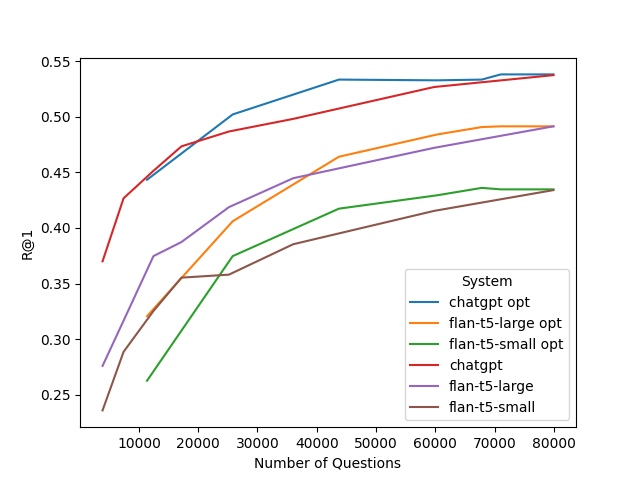}
            \caption{R@1}
            \label{fig:subfig1}
        \end{subfigure} &
        \begin{subfigure}[b]{0.33\textwidth}
            \centering
            \includegraphics[width=\textwidth]{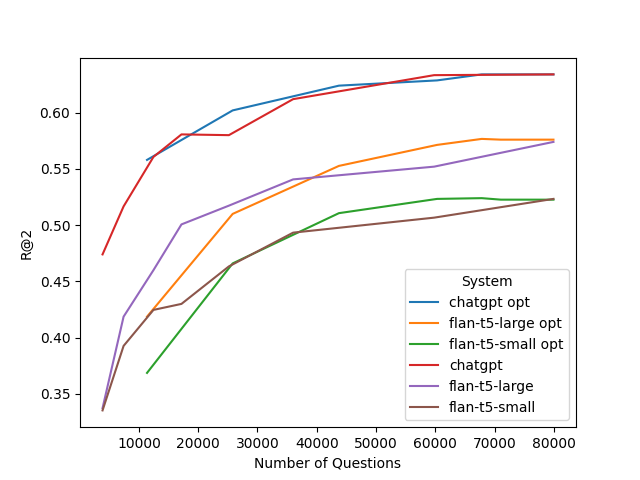}
            \caption{R@2}
            \label{fig:subfig2}
        \end{subfigure}
        \begin{subfigure}[b]{0.33\textwidth}
            \centering
            \includegraphics[width=\textwidth]{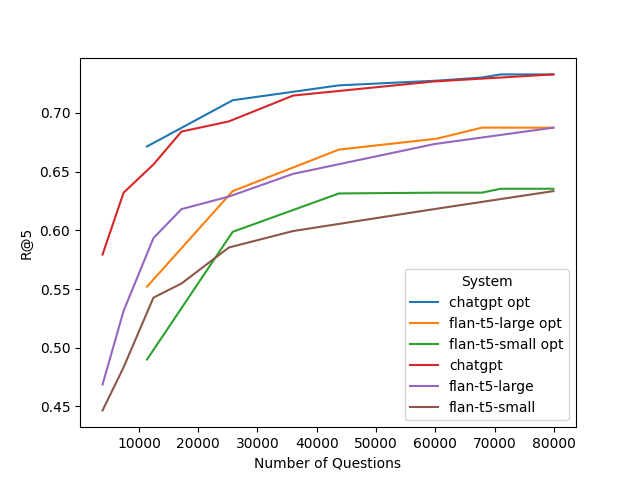}
            \caption{R@5}
            \label{fig:subfig3}
        \end{subfigure}
    \end{tabular}
    \caption{Comparing question generation systems using retrieval on BiPaR with all-mpnet-base-v2 embedder and including optimal question selection.}
    \label{fig:efficiency_optimal}
\end{figure*}

\begin{table*}[htbp!]
\centering
\small 
\begin{tabular}{l|ccc|ccc}
\toprule
& \multicolumn{3}{c}{Retrieval - random} & \multicolumn{3}{|c}{Retrieval - pruned}  \\
System & R@1 nAUC & R@2 nAUC & R@5 nAUC & R@1 nAUC & R@2 nAUC & R@5 nAUC \\
\midrule
chatgpt-3.5 & 0.474 & 0.574 & 0.670 & 0.444 & 0.528 & 0.616 \\
flan-t5-large & 0.414 & 0.500 & 0.610 & 0.382 & 0.461 & 0.561 \\
flan-t5-base & 0.370 & 0.460 & 0.572 & 0.349 & 0.426 & 0.531 \\
flan-t5-small & 0.363 & 0.455 & 0.562 & 0.341 & 0.421 & 0.523 \\
   \bottomrule
    \end{tabular}
\caption{Comparison of question generation systems applied to contexts from BiPaR using all-mpnet-base-v2. }
\label{tab:comp_eval}
\end{table*}

\begin{figure*}[htbp!]
    \centering
    \begin{tabular}{cc}
        \begin{subfigure}[b]{0.33\textwidth}
            \centering
            \includegraphics[width=\textwidth]{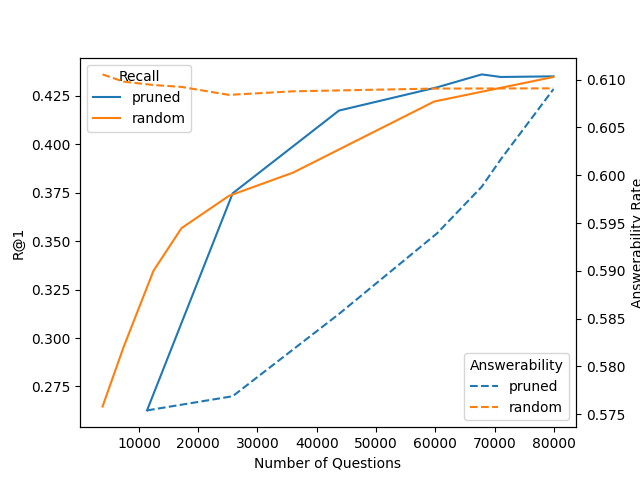}
            \caption{R@1}
            \label{fig:subfig1}
        \end{subfigure} &
        \begin{subfigure}[b]{0.33\textwidth}
            \centering
            \includegraphics[width=\textwidth]{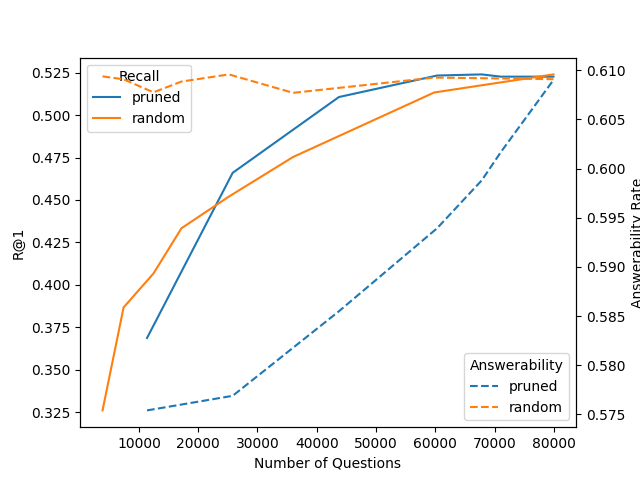}
            \caption{R@2}
            \label{fig:subfig2}
        \end{subfigure}
        \begin{subfigure}[b]{0.33\textwidth}
            \centering
            \includegraphics[width=\textwidth]{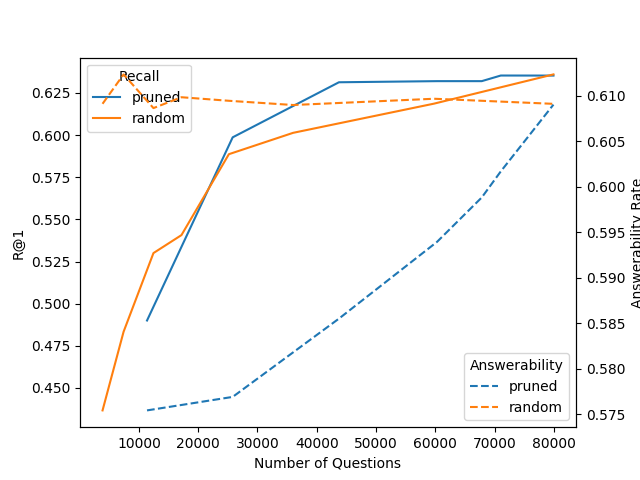}
            \caption{R@5}
            \label{fig:subfig3}
        \end{subfigure}
    \end{tabular}
    \caption{Answerability rates for optimal (pruned) and random lines for specifically flan-t5-small as the question generation system.}
    \label{fig:ans_rates}
\end{figure*}

\section{Additional Results}

\subsection{Open-source question generation systems}
\label{sec:app:open}

Figure \ref{fig:efficiency} is presented using ChatGPT as the question generation system over the unstructured atoms. Here, we extend the results to explore the behaviour of generating questions over structured atoms from the BiPaR dataset using open-source large language models for the question generation systems. We focus on the all-mpnet-base-v2 as the embedding system for retrieval.

The plots of the randomly selected questions and the corresponding optimal lines is presented in Figure \ref{fig:efficiency_optimal}. 
Here, the Flan-T5 \citep{chung2024scaling} model series is selected as open-source models for question generation.
It is clear that for the selected open-source models, the optimal lines envelope the randomly selected questions in a manner similar to the closed-source ChatGPT model. However, we do note that with a small sample of questions, the randomly selected set of questions outperforms the optimally diverse set for the Flan-T5 models. See Section \ref{sec:app:unans} for the justification for this observation.

Table \ref{tab:comp_eval} provides the summary statistics for the normalized area under each of these curves (nAUC) where the x axis is scaled to be between 0 and 1.

\subsection{Unanswerability analysis}
\label{sec:app:unans}

A potential concern of the generated questions from a given question generation system is that we assume the question is appropriate for the atom on which it was generated. A form of appropriateness is captured by the \textit{unanswerability} of the question. We aim to measure the unanswerability of the generated questions to understand to what degree they are appropriate.

SQuAD 2.0 \citep{rajpurkar2018know} is annotated with answerable and unanswerable questions over reading comprehension contexts. Hence, we use the validation split of this dataset to assess a zero-shot Flan-T5-Large as an unanswerability system. SQuAD 2.0 validation split consists of 5,928 answerable questions and 5,945 unanswerable questions. There are 2,067 context paragraphs in total. 

The system is prompted to return \textit{yes} if a question is unanswerable and \textit{no} if unanswerable. As is common with instruction-tuned models for classification tasks, a binary probability distribution is formed by applying Softmax to the logits associated with the \textit{yes} and \textit{no} tokens from the token vocabulary of the model. This system is able to achieve an F1 score of 86.5 with a precision and recall of 83.0 and 90.3 respectively.

Therefore, Figure \ref{fig:ans_rates} presents the answerability rates for the optimal and random lines for the different recall rates using the Flan-T5-Small system. It is clear that the answerability of the set of questions for the optimal set (referred to here as pruned) drops dramatically with fewer questions. This is somewhat expected because the optimal set of questions are selected to be as diverse as possible from each other. Thus, it is more likely that obscure (unanswerable) questions are selected from the pool of generated questions if diversity is the criteria for optimization.

\section{Licenses}

SQuAD is shared under the attribution-sharealike 4.0 international (CC BY-SA 4.0) license. BiPaR is shared under the attribution-noncommercial 4.0 international (CC BY-NC 4.0) license. CLAPNQ is shared under the Apache-2.0 license.

\end{document}